\documentclass[a4paper,fleqn]{cas-sc}

\usepackage{amssymb}
\usepackage{longtable}
\usepackage{amsmath}
\usepackage{graphicx}%
\usepackage{multirow}%
\usepackage{amsmath,amssymb,amsfonts}%
\usepackage{amsthm}%
\usepackage{mathrsfs}%
\usepackage[title]{appendix}%
\usepackage{xcolor}%
\usepackage{textcomp}%
\usepackage{manyfoot}%
\usepackage{booktabs}%
\usepackage{algorithm}%
\usepackage{algorithmicx}%
\usepackage{algpseudocode}%
\usepackage{listings}%
\usepackage{lscape}  
\usepackage{textalpha} 
\usepackage{fontawesome}
\usepackage{array}
\usepackage{url}
\usepackage[normalem]{ulem}
\usepackage[authoryear,longnamesfirst]{natbib}

\usepackage{float}

\def\tsc#1{\csdef{#1}{\textsc{\lowercase{#1}}\xspace}}
\tsc{WGM}
\tsc{QE}

\begin{document}
\setcounter{topnumber}{5}
\setcounter{bottomnumber}{5}
\setcounter{totalnumber}{10}
\renewcommand{\topfraction}{0.9}
\renewcommand{\bottomfraction}{0.9}
\renewcommand{\textfraction}{0.1}
\renewcommand{\floatpagefraction}{0.8}

\shorttitle{\textit{Verify as You Go}}    

\shortauthors{Sallami \textit{et al.}}  

\title{\textit{Verify as You Go}: An LLM-Powered Browser Extension for Fake News Detection}

\author[inst1]{Dorsaf Sallami}

\author[inst1]{Esma Aïmeur}

\affiliation[inst1]{organization={Department of Computer Science and Operations Research, University of Montreal},
            addressline={André-Aisenstadt Pavilion, 2920, chemin de la Tour}, 
            city={Montreal},
            postcode={H3T 1J4}, 
            state={Quebec},
            country={Canada}}

\begin{abstract}
The rampant spread of fake news in the digital age poses serious risks to public trust and democratic institutions, underscoring the need for effective, transparent, and user-centered detection tools. Existing browser extensions often fall short due to opaque model behavior, limited explanatory support, and a lack of meaningful user engagement. This paper introduces \textit{Aletheia}, a novel browser extension that leverages Retrieval-Augmented Generation (RAG) and Large Language Models (LLMs) to detect fake news and provide evidence-based explanations. \textit{Aletheia} further includes two interactive components: a \textit{Discussion Hub} that enables user dialogue around flagged content and a \textit{Stay Informed} feature that surfaces recent fact-checks. Through extensive experiments, we show that \textit{Aletheia} outperforms state-of-the-art baselines in detection performance. Complementing this empirical evaluation, a complementary user study with 250 participants confirms the system’s usability and perceived effectiveness, highlighting its potential as a transparent tool for combating online fake news.
\end{abstract}




\maketitle


\section{Introduction}\label{sec1}
Fake news is more than just falsehood; it’s a powerful tool designed to deceive and mislead, often spreading at an alarming pace through the digital landscape~\citep{sallami2025exploring}. Its unchecked proliferation has deepened societal divisions and weakened democratic foundations, raising urgent concerns among researchers, policymakers, and the public~\citep{balakrishnan2022infodemic,sallami2023hype}.

Efforts to combat fake news have increasingly engaged researchers and digital platform operators. The predominant approach involves leveraging Artificial Intelligence (AI) models for fake news detection~\citep{sallami2023trust,amri2021exmulf}. While these models demonstrate notable advancements, they often fall short of delivering comprehensive, user-centric solutions for effectively identifying fake news. A discernible gap persists between state-of-the-art machine learning methodologies and practical tools designed for end-users. Although AI-based detection systems continue to evolve, there is a critical need for a holistic approach that not only ensures accuracy but also prioritizes accessibility and user engagement. Integrating Machine Learning (ML) algorithms into browser extensions offers a direct mechanism for addressing these challenges~\citep{sallami2025exploring}.

Browser extensions have become a promising means of integrating advanced computational methods with accessible, user-oriented detection tools. By seamlessly integrating ML capabilities into users' browsing experiences, these extensions facilitate the identification of fake news in real-time with minimal friction. However, current browser-based solutions demonstrate significant limitations, as illustrated in Table~\ref{tab:comparison}. These include a lack of transparency regarding their underlying algorithms~\citep{sallami2025explain}, reliance on conventional machine learning models with limited adaptability, presentation of detection results without explanatory context, and a predominant focus on classification tasks rather than providing complementary functionalities that support informed user decision-making.

To address these limitations, this research introduces \textit{Aletheia}\footnote{The name \textit{Aletheia} (\textalpha\textlambda\texteta\texttheta\textepsilon\textiota\textalpha) is an ancient Greek word commonly translated as "truth"~\citep{wolenski2004aletheia}. The name reflects the tool’s purpose of exposing misinformation, aligning with its philosophical roots in truth-seeking.}, an innovative browser extension designed to enhance fake news mitigation. The proposed solution advances existing methodologies by:
\begin{enumerate}
    \item Integrating Retrieval-Augmented Generation (RAG) and Large Language Models (LLMs) to detect fake news while providing evidence-based explanations derived from RAG-powered Google searches.
    \item Incorporating a Discussion Hub that enables users to post and comment on suspected misinformation instances, fostering community-driven analysis and engagement.
    \item Implementing a Stay Informed feature that delivers real-time fact-checking updates, ensuring users have access to the latest verified information.
\end{enumerate}

We organize the remainder of this paper as follows: Section~\ref{sec1} reviews the related work. Section~\ref{sec2} presents the overall architecture and core components of our proposed browser extension. Section~\ref{sec:modelperformance} reports the experimental evaluations and model-based analyses that benchmark detection performance against state-of-the-art baselines, followed by Section~\ref{sec:userstudy}, which details a user study evaluating usability and perceived utility. Finally, Section~\ref{secConclusion} concludes the paper with a summary of our contributions and potential directions for future research.

\section{Related Works}\label{sec2}
In this section, we review prior efforts relevant to our work. First, we explore evidence-based methods, both classical and LLM-driven. Second, we examine browser extensions designed for fake news detection. 
\subsection{Evidence-Based Fake News Detection}\label{sec:baselines}
\subsubsection{Classical Evidence-Based Approaches}
Several models introduced increasingly more sophisticated evidence-aware mechanisms. DeClarE~\citep{popat2018declare} employs a Bidirectional Long Short-Term Memory network to capture the semantic nuances of evidence, using an attention-based interaction mechanism to compute relevance scores between claims and supporting texts. Building upon this foundation, HAN~\citep{ma2019sentence} integrates Gated Recurrent Unit embeddings with two dedicated modules, one for topic consistency and the other for semantic entailment, both employing sentence-level attention to simulate nuanced interactions between claims and evidence. Extending the interpretability of verification systems, EHIAN~\citep{wu2021evidence} introduces an evidence-aware hierarchical interactive attention network that guides the selection of semantically relevant and plausible evidence. MAC~\citep{vo2021hierarchical} further enhances interpretability by combining multi-head word-level and document-level attention, allowing for hierarchical analysis of fake news at both lexical and evidential layers. Meanwhile, GET~\citep{xu2022evidence} models claims and evidence as graph-structured data, capturing intricate semantic relationships and mitigating information redundancy through a semantic structure refinement layer. In a complementary direction, MUSER~\citep{liao2023muser} improves verification performance by adopting a multi-step evidence retrieval strategy that capitalizes on the interdependencies among multiple evidence fragments. Finally, ReRead~\citep{hu2023read} focuses on real-world document retrieval guided by criteria of plausibility, sufficiency, and relevance, refining the evidence selection process to support robust claim validation. Despite strong performance on curated benchmarks, these classical approaches typically rely on static evidence pools and are therefore limited in real-time and evolving fake news scenarios.
\subsubsection{LLM-Based Approaches}
Recent advances in LLMs have transformed fake news detection by enabling contextual understanding, few-shot learning, and complex reasoning without task-specific fine-tuning~\citep{sallami2024deception,xu2024comparative, sallami2025claimveragents}. For instance, Vicuna-7B~\citep{chiang2023vicuna}, an open-source model fine-tuned from LLaMA using ShareGPT data, has gained popularity for its strong performance in conversational tasks. WEBGLM~\citep{liu2023webglm} builds on the GLM framework, enhancing question answering by integrating real-time web search. ProgramFC~\citep{pan2023fact} introduces a modular fact-checking approach by breaking down complex claims into simpler subclaims and retrieving evidence using Codex; we adopt its open-book configuration. STEEL~\citep{li2024re} further advances the field with a retrieval-augmented framework that automatically identifies and extracts relevant web evidence for claim verification.

Khaliq \textit{et al.}~\citeyear{khaliq2024ragar} investigated the application of RAG for multimodal political fact-checking, wherein textual claims are evaluated in conjunction with corresponding visual evidence. In addition, they introduced the Chain of RAG (CoRAG) and Tree of RAG (TreeRAG) frameworks to enhance the reasoning capabilities and interpretability of RAG-based systems~\citep{khaliq2024ragar}. Similarly, Markey \textit{et al.}~\citeyear{markey2025rags} employed RAG to support the drafting of medical documentation from LLM outputs, demonstrating that the integration of retrieval components improves the overall writing quality and ensures the inclusion of current, evidence-based information~\citep{markey2025rags}.

RAG, as initially proposed by Lewis \textit{et al.}~\citeyear{lewis2020retrieval} and further advanced by Ram \textit{et al.}~\citeyear{ram2023context}, enables the adaptation of LLMs to automated fact-checking tasks by incorporating externally retrieved evidence into the generative process~\cite{pan2023qacheck, chen2023complex, zhang2024reinforcement}. Nevertheless, the reliability of retrieved evidence remains a critical concern, as sources of uncertain credibility may introduce misinformation, thereby generating conflicting or misleading evidence~\cite{hong2023so}.

While RAG has improved LLM performance on general tasks~\citep{gao2023retrieval,wang2024large}, its use in fake news detection remains limited. Few systems combine RAG with real-time evidence retrieval, and even fewer support multi-step, iterative verification. 
\subsection{Fake News Detection Browser Extensions}
Empirical studies reveal that individuals frequently disseminate misinformation without prior verification, often acting on impulse~\citep{marsili2021retweeting}. This behavioral tendency underscores the necessity for mechanisms that support users in evaluating the credibility of online content \citep{geeng2020fake}. In response, research has extensively examined the efficacy of cognitive interventions, such as nudges, designed to prompt reflection before sharing information. Findings indicate that these interventions effectively mitigate the spread of falsehoods \citep{pennycook2020fighting, marsili2021retweeting, gwebu2022can} and enhance individuals’ capacity to discern misleading headlines, particularly when accompanied by contextual explanations \citep{aimeur2025too, kirchner2020countering}.

A variety of browser extensions have emerged as essential tools in the fight against fake news, significantly enhancing users' ability to assess the reliability of online content. Notable examples available in the Google Chrome Extension Store include The Fact Checker~\citep{FactChecker} and Media Bias Fact Check~\citep{MediaBiasFactCheck}. However, their methodologies remain opaque, as no peer-reviewed research or technical documentation clarifies their underlying models, raising concerns about reproducibility and efficacy. Alternative approaches typically involve flagging content from known fake news sources using continuously updated lists, as demonstrated by extensions like B.S. Detector~\citep{BSDetector}, which poses practical challenges due to the frequent need for list updates. Similarly, tools like FactIt \citep{velasco2023factit} rely on traditional machine learning techniques such as logistic regression—pre-dating advances in LLMs—and frame fake news detection as a simplistic binary classification task. These methods, however, struggle to adapt to the evolving complexity of misinformation, yielding marginal accuracy improvements \citep{kuntur2024under}.

More advanced solutions, including Check-It~\citep{paschalides2021check}, TrustyTweet~\citep{hartwig2019trustytweet}, BRENDA~\citep{botnevik2020brenda}, and ShareAware~\citep{von2020nudging}, leverage deep neural networks and machine learning algorithms for fake news detection. Despite these advancements, our review reveals that no existing browser extension currently employs LLMs for this purpose, even though LLMs surpass these models in detection accuracy~\citep{kuntur2024under}. This gap is clearly reflected in the comparative analysis presented in Table~\ref{tab:comparison}.

Another critical limitation of current solutions is their lack of explainability. While some tools, such as COVID-FakeExplainer~\citep{warman2023covidfakeexplainer}, incorporate SHAP (SHapley Additive exPlanations)~\citep{lundberg2017unified} to justify predictions, these technical explanations often alienate non-expert users. This disconnect between algorithmic outputs and user comprehension undermines trust and practical utility~\citep{epstein2022explanations}, highlighting the need for intuitive, human-centric interfaces that align with how users process information \citep{gong2024integrating}. Furthermore, existing extensions focus narrowly on detection, offering no supplementary features to enhance user engagement. Our work addresses this gap by introducing a browser extension that integrates LLM-driven detection with two novel components: (1) a discussion hub fostering user dialogue about disputed content and (2) real-time updates on the latest fact-checking outcomes. 
\begin{landscape}
\begin{table}[]
\centering
\caption{Comparison of Browser Extensions for Fake News Detection. 
\newline \textbf{Legend:} 
\faCheck = Feature present or supported; 
\faTimes = Feature absent or unsupported; 
\faQuestionCircle = Unknown or undocumented; 
\textsuperscript{\faBan} = Not peer-reviewed.
\faStar = UX feedback from Google Reviews
 available.}
\label{tab:comparison}
\begin{tabular}{
|p{3.5cm}                                  
|>{\centering\arraybackslash}m{3cm}        
|>{\centering\arraybackslash}m{2.5cm}      
|>{\centering\arraybackslash}m{4cm}       
|>{\centering\arraybackslash}m{3cm}        
|>{\centering\arraybackslash}m{3cm}        
|>{\centering\arraybackslash}m{2cm}|       
}

\hline
\multicolumn{1}{|>{\centering\arraybackslash}p{3.5cm}|}{\textbf{Extension}} & \textbf{Detection Model} & \textbf{Explanation} & \textbf{Community Engagement} & \textbf{Fact Updates} & \textbf{UX Evaluation} \\
\hline
The Fact Checker \textsuperscript{\faBan} & \faQuestionCircle  & \faTimes & \faTimes & \faTimes & \faStar \\
\hline
Media Bias Fact Check \textsuperscript{\faBan}	 & \faQuestionCircle  & \faTimes & \faTimes & \faTimes & \faStar \\
\hline
B.S. Detector \textsuperscript{\faBan}	& \faQuestionCircle  & \faTimes & \faTimes & \faTimes & \faStar \\
\hline
FactIt \citep{velasco2023factit} & Logistic Regression  & \faTimes & \faTimes & \faTimes & \faTimes \\
\hline
Check-It \citep{paschalides2021check} & Logistic Regression  & \faTimes	 & \faTimes & \faTimes & \faTimes \\
\hline
TrustyTweet \citep{hartwig2019trustytweet} & Rule-based indicators  & User-visible cues	 & \faTimes & \faTimes & \faCheck (27 participants)\\
\hline
BRENDA \citep{botnevik2020brenda} & Neural network  & Highlights claim	 & \faTimes &  \faTimes & \faTimes \\
\hline
COVID-FakeExplainer \citep{warman2023covidfakeexplainer} & BERT    & SHAP  & \faTimes	 & \faTimes & \faTimes \\
\hline
 FeedReflect \citep{bhuiyan2018feedreflect}&  Source-based & \faTimes  & \faTimes	 & \faTimes & \faCheck (16 participants) \\
\hline
ShareAware \citep{von2020nudging}& Link Analysis   & \faTimes  & \faTimes	 & \faTimes & \faTimes \\
\hline
FNDaaS \citep{papadopoulos2022fndaas}& Random Forest   & \faTimes  & \faTimes	 & \faTimes & \faTimes \\
\hline
\textbf{\textit{Aletheia}} & LLM + RAG  & Natural-language, evidence-based & \faCheck	 & \faCheck & \faCheck (250 participants) \\
\hline
\end{tabular}
\end{table}
\end{landscape}
\section{Methodology}\label{sec2}
\begin{figure}[]
\centering
\includegraphics[width=0.8\textwidth]{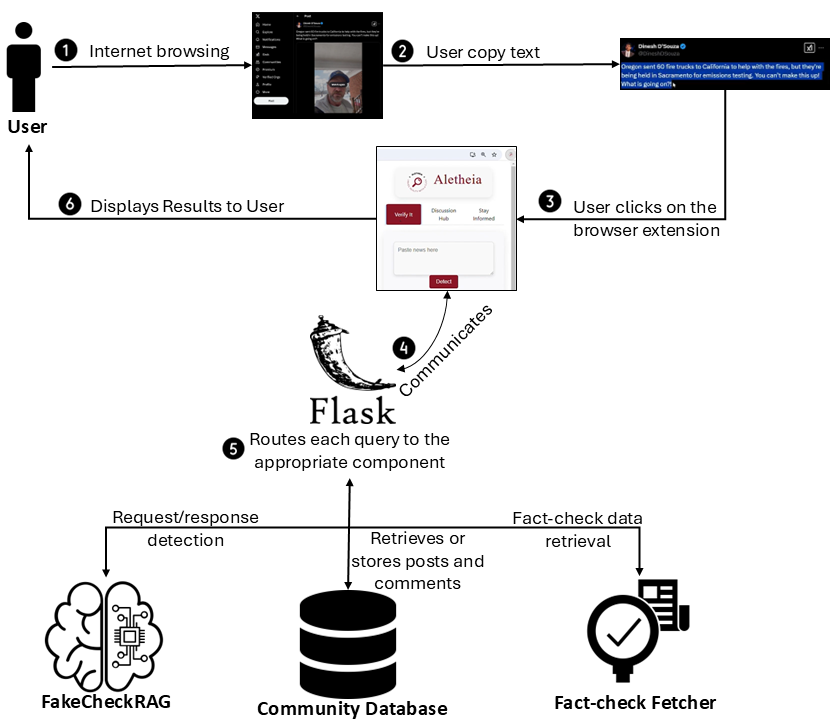}
\caption{ \textit{Aletheia} system architecture. The client-side browser extension interfaces with a Flask-based backend server. The backend comprises three primary modules: Fact-Check Fetcher, Community Database, and FakeCheckRAG.}
\label{fig:architecture}
\end{figure}

 \textit{Aletheia} is a modular fact-checking system that integrates interactive user input with automated retrieval-augmented verification. The architecture comprises two main layers: a browser-based user interface, a server-side backend. Figure~\ref{fig:architecture} provides an overview of the system architecture.

\subsection{Frontend: Browser Extension}
The frontend of \textit{Aletheia} is implemented as a browser extension compatible with Google Chrome. When the user clicks the extension icon to initiate the backend.

\begin{figure}
\centering
\includegraphics[width=0.7\textwidth]{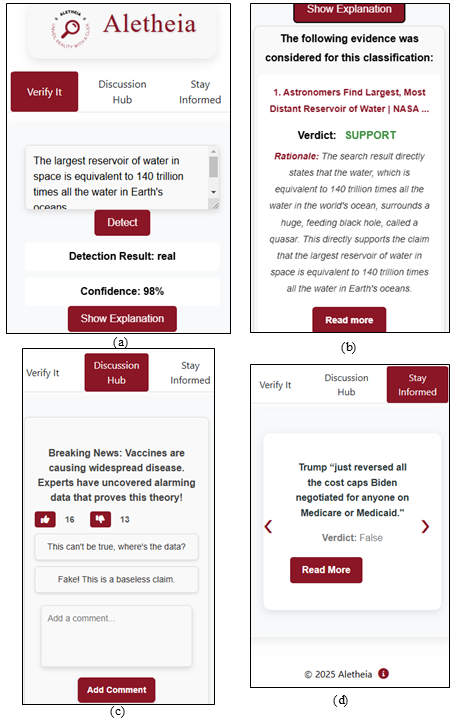}
\caption{Snapshots of \textit{Aletheia}’s user interface: (a) claim submission, (b) system explanation, (c) community discussion, and (d) recent fact-checks.}
\label{fig:screenshots}
\end{figure}

Users interact with \textit{Aletheia} through three main components. The claim verification interface allows users to input statements to be fact-checked. After detection, the system returns a verdict along with evidence-based explanations. The explanation panel provides a natural-language rationale for the system’s output. The discussion hub fosters community interaction by allowing comments and voting on contentious claims. Lastly, a newsfeed section presents up-to-date fact-checks aggregated from verified external sources. Figure~\ref{fig:screenshots} illustrates these functionalities.

\subsection{Backend: Server Architecture}
The backend is implemented using the Flask web framework and exposes a suite of RESTful APIs\footnote{RESTful APIs (Representational State Transfer Application Programming Interfaces) are web services that follow REST principles, allowing clients to access and manipulate resources over HTTP using standard methods like GET, POST, PUT, and DELETE.} to handle frontend requests and coordinate various processing modules. It consists of three major subsystems: Fact-Check
Fetcher, Community Database and FakeCheckRAG.

The Fact-Check Fetcher module connects to the Google Fact Check Tools API to obtain recent verdicts on publicly reviewed claims. Claims are filtered and parsed to ensure freshness and structural consistency. The Community Database supports discussion and interaction by managing persistent data on posts, comments, and user votes using PostgreSQL\footnote{PostgreSQL is an open-source relational database management system known for its robustness, extensibility, and compliance with SQL standards.}. The most computationally intensive component is FakeCheckRAG, a retrieval-augmented reasoning engine that performs claim verification using web evidence and LLMs.

\subsubsection{FakeCheckRAG: Retrieval-Augmented Verification Engine}
\begin{figure}[]
\centering
\includegraphics[width=\textwidth]{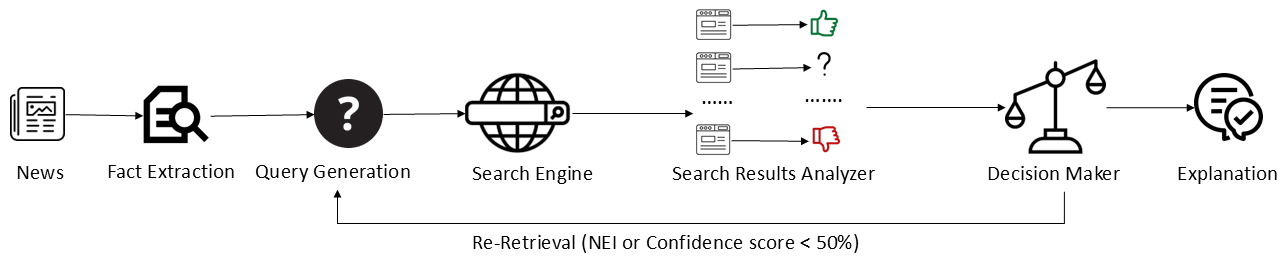}
\caption{The FakeCheckRAG pipeline from claim parsing to classification.}
\label{fig:RAG}
\end{figure}

FakeCheckRAG is the central reasoning module within \textit{Aletheia}’s backend, responsible for assessing the truthfulness of user-submitted claims. The system implements a tailored Retrieval-Augmented Generation pipeline for fact verification that replaces static knowledge bases with web retrieval and pairs retrieval with verification-oriented reasoning. Our approach extends standard RAG by (1) \textit{dynamic retrieval} from live web sources instead of static knowledge bases, (2) \textit{verification-oriented reasoning} using evidence classification rather than free-form generation, and (3) iterative refinement with query reformulation. This specialized approach addresses the unique challenges of real-time claim verification on evolving web content. It combines structured web retrieval, source filtering, LLM-based textual inference, and iterative re-querying to produce classification results with interpretability and confidence scores. The pipeline is shown in Figure~\ref{fig:RAG}.

Upon receiving a claim \( C \), the system extracts its primary fact. A structured search query \( Q \) is automatically generated and submitted to the Google Search API\footnote{\url{https://developers.google.com/custom-search/v1/overview}}. In our configuration we use \texttt{GPT-4} for key-claim extraction and query generation, retrieve the top ten web results (title/URL/snippet) per query. The set of retrieved URLs \( U \) is then filtered by removing domains known for fake news, resulting in a cleaned subset \( U' \), based on a curated blacklist of 1,044 unreliable sources~\citep{papadogiannakis2023funds}.

Each remaining source \( u \in U' \) is then analyzed by an LLM to determine its relationship to the original claim \( C \), producing a label \( r \in \{\text{Support}, \text{Refute}, \text{Unrelated}\} \). This process yields a set of textual evidence passages, denoted as \( \mathcal{E} \). These LLM-based inference calls are executed in parallel, using a throttled request scheduler to maintain compliance with API rate limits and ensure system responsiveness. The collection of evidence-level results is stored as \( \mathcal{R} \).

The system then aggregates the results \( \mathcal{R} \) to derive a final classification label \( L \in \{\text{Real}, \text{Fake}, \text{NEI}\} \) (Not Enough Information). The aggregation is performed with a separate \texttt{GPT-4} call over the per-source labels/rationales. To ensure reliability, each classification is assigned a confidence score (0–100\%), mitigating inconsistencies~\citep{xiong2023can} and hallucinations~\citep{ye2022unreliability}. In parallel, a natural-language explanation \( e \) is generated to summarize the justification behind the verdict.

If the outcome is \( L = \text{NEI} \) or the confidence score \( s \) falls below a threshold \( \tau \), a re-retrieval phase is initiated. Concretely, we set \(\tau{=}50\) and allow up to \(\textit{max\_iters}{=}3\) rounds: on each round we regenerate the key claim and query, repeat retrieval and judging, and terminate early once \(s\ge \tau\). The system reformulates the original query into a new variant \( Q' \), performs a second round of retrieval, and updates the evidence set \( \mathcal{E} \).  This loop proceeds for up to \( max\_iters = 3 \) rounds, based on results from the ablation study in Section~\ref{sec:ablation}, or terminates earlier if a confident verdict is reached. This full pipeline is formally described in Algorithm~\ref{alg:fakecheckrag}.

\begin{algorithm}[h!]
\caption{FakeCheckRAG}
\label{alg:fakecheckrag}
\begin{algorithmic}[1]
\Require Claim $C$, Confidence threshold $\tau$, Maximum iterations $max\_iters \gets 3$
\Ensure Label $L \in \{\text{Real}, \text{Fake}, \text{NEI}\}$, Confidence score $s \in [0, 100]$, Explanation $e$

\State $Q \gets \textsc{GenerateSearchQuery}(C)$
\State $i \gets 0$
\State $\mathcal{E} \gets \emptyset$

\While{$i < max\_iters$}
    \State $U \gets \textsc{SearchAPI}(Q)$
    \State $U' \gets \textsc{FilterSources}(U)$ \Comment{Filter low-credibility sources}

    \ForAll{$u \in U'$}
        \State $t \gets \textsc{ExtractRelevantText}(u)$
        \State $\mathcal{E} \gets \mathcal{E} \cup \{t\}$
    \EndFor

    \State $\mathcal{R} \gets \emptyset$
    \ForAll{$e \in \mathcal{E}$ \textbf{in parallel}}
        \State $r \gets \textsc{LLMInfer}(C, e)$
        \State $\mathcal{R} \gets \mathcal{R} \cup \{r\}$
    \EndFor

    \State $L \gets \textsc{AggregateLabels}(\mathcal{R})$
    \State $s \gets \textsc{ComputeConfidence}(\mathcal{R})$
    \State $e \gets \textsc{GenerateExplanation}(\mathcal{R}, \mathcal{E})$

    \If{$L \ne \text{NEI}$ \textbf{and} $s \ge \tau$}
        \State \Return $(L, s, e)$ \Comment{Confident decision reached}
    \EndIf

    \State $Q \gets \textsc{ReformulateQuery}(\mathcal{E})$
    \State $i \gets i + 1$
\EndWhile

\State \Return $(L, s, e)$ \Comment{Return result from final iteration}
\end{algorithmic}
\end{algorithm}

\section{Model Performance Evaluation}\label{sec:modelperformance}
This evaluation aims to quantify the accuracy of FakeCheckRAG. 
\subsection{Experimental Setup}
\subsubsection{Datasets}
To evaluate the performance of \textit{FakeCheckRAG}, we conducted extensive experiments on two real-world datasets, including two English-language sources: LIAR and PolitiFact. The LIAR dataset comprises 9,252 real news instances and 3,555 fake ones, totaling 12,807 samples. PolitiFact contains 399 real and 345 fake news items, amounting to 744 in total.
\subsubsection{Baselines}
To assess the performance of \textit{FakeCheckRAG}, we compare it against a comprehensive set of baselines, including \textbf{classical evidence-based approaches}, widely used and competitive static baselines, as well as \textbf{LLM-based methods}, encompassing both retrieval and non-retrieval variants:

\begin{enumerate}
    \item \textbf{Classical evidence-based methods}\\
  \begin{itemize}
\item DeClarE~\citep{popat2018declare} utilizes a Bidirectional Long Short-Term Memory network to capture the semantic representations of evidence, computing evidence scores through an attention-based interaction mechanism.

\item HAN~\citep{ma2019sentence} employs Gated Recurrent Unit embeddings and integrates two specialized modules: one for assessing topic consistency and another for evaluating semantic entailment. Both modules leverage a sentence-level attention mechanism to effectively simulate the interaction between claims and evidence.
 
\item EHIAN~\citep{wu2021evidence} achieves interpretable claim verification through an evidence-aware hierarchical interactive attention network, which facilitates the exploration of more plausible and semantically relevant evidence.
 
\item MAC~\citep{vo2021hierarchical} combines multi-head word-level attention with multi-head document-level attention, enabling interpretable fake news detection at both the lexical (word-level) and evidential (document-level) granularity.
 
 \item GET~\citep{xu2022evidence} models claims and associated evidence as graph-structured data, enabling the exploration of complex semantic relationships. Additionally, it addresses information redundancy by incorporating a semantic structure refinement layer.
 
 \item MUSER~\citep{liao2023muser} adopts a multi-step evidence retrieval strategy, leveraging the interdependencies among multiple pieces of evidence to enhance verification performance.
 
 \item ReRead~\citep{hu2023read} retrieves relevant evidence from real-world documents by applying the standards of plausibility, sufficiency, and relevance.
\end{itemize}

    \item \textbf{LLM-based methods}\\
   \begin{itemize}
\item GPT-3.5-turbo~\citep{GPT-3.5-turbo} is a sibling model to InstructGPT, sharing its ability to comprehend and respond to prompts with detailed explanations. In this work, we employ gpt-3.5-turbo as the baseline model.

\item Vicuna-7B~\citep{chiang2023vicuna}  is a widely adopted, fully open-source base model for LLMs, developed by fine-tuning LLaMA on user-contributed conversational data collected from ShareGPT.

\item WEBGLM~\citep{liu2023webglm} is a web-enhanced question-answering system built on the WebGPT model~\citep{nakano2021webgpt}. The model retrieves relevant content from the Internet and feeds it into LLMs for analysis. The 2B variant with Bing search integration is utilized.

\item ProgramFC~\citep{pan2023fact} is a fact-checking model that decomposes complex claims into simpler subclaims, which are addressed using a shared library of specialized functions. The model employs strategic retrieval powered by Codex to perform fact-checking. 

\item STEEL~\citep{li2024re} adopts a retrieval-augmented LLM framework, leveraging GPT-3.5-turbo to automatically and extract key evidence from web sources for claim verification.
\end{itemize}
\end{enumerate}

\subsubsection{Implementation details}
Since our model operates without the need for training data, the entire dataset is used exclusively for testing. This evaluation strategy is consistently applied across all datasets. We employ OpenAI GPT-4 for all reasoning tasks. For the LLM-based baselines, we set the temperature to 0 and employ standardized prompt templates, as summarized in Table~\ref{tab:agent-prompts}. Hyperparameters for baseline methods are adopted from their respective original works, with key parameters carefully tuned to ensure optimal performance. We formulate fake news detection as a binary classification task, and report results using F1 score (F1), Precision (P), and Recall (R) as evaluation metrics.
\subsection{Results}
We compare the performance of our proposed model, \textit{FakeCheckRAG}, with eleven existing baselines, comprising seven evidence-based and four LLM-based methods. Tables~\ref{tab:results1} and~\ref{tab:results2} present detailed results on the PolitiFact and LIAR datasets, respectively. 
\subsubsection{Overall Results}
\textit{FakeCheckRAG} consistently outperforms all baselines across both the PolitiFact and LIAR datasets, demonstrating the model’s superior detection capabilities.

Classical methods employ static evidence retrieval mechanisms. Among these, MUSER exhibits the strongest performance, with F1 scores of 0.75 for real claims and 0.70 for fake claims on the PolitiFact dataset, and 0.64 for both real and fake claims on LIAR. However, these methods are consistently outperformed by \textit{FakeCheckRAG}, which achieves F1 scores of 0.85 (real) and 0.83 (fake) on PolitiFact, and 0.87 (real) and 0.83 (fake) on LIAR. These results indicate a notable performance improvement over classical systems. This performance gap reflects the limitations of earlier architectures in integrating evidence contextually and reasoning over claims with the depth required for high-accuracy fact-checking.

LLM-based baselines leverage the generative capabilities of modern language models. Among them, STEEL delivers the strongest results, obtaining F1 scores of 0.78 (real) and 0.72 (fake) on PolitiFact, and 0.68 (real) and 0.74 (fake) on LIAR. Nevertheless, \textit{FakeCheckRAG} significantly outperforms these models. Compared to STEEL, our system achieves improvements of 7 F1 points on real claims and 11 points on fake claims on the PolitiFact dataset, and 19 points on real claims and 9 points on fake claims on LIAR. These findings demonstrate the advantages of our architecture, which integrates retrieval-augmented generation with GPT-4 and incorporates a re-search mechanism that enables iterative query refinement.

\begin{table}[htpb]
\centering
\footnotesize
\caption{
Comparison of our model's performance against baselines on PolitiFact dataset.}\label{tab:results1}
\begin{tabular}{llllllll}
\hline
\multicolumn{2}{l}{\multirow{2}{*}{Method}}       & \multicolumn{3}{c}{Real}                      & \multicolumn{3}{c}{Fake}                      \\ \cline{3-8} 
\multicolumn{2}{l}{}                                                                                                                               & F1            & P             & R             & F1            & P             & R             \\ \hline
\multirow{7}{*}{\begin{tabular}[c]{@{}l@{}}Classical\\  evidence\\ based\\  methods\end{tabular}} & DeClarE                                        & 0.65          & 0.68          & 0.67          & 0.65          & 0.61          & 0.66          \\  
  & HAN & 0.67  & 0.67   & 0.68  & 0.64   & 0.65  & 0.63  \\ 
                                             & EHIAN   & 0.67   & 0.68   & 0.65  & 0.65   & 0.62  & 0.62  \\ 
                                              & MAC    & 0.70          & 0.69          & 0.70          & 0.65          & 0.65          & 0.64          \\   & GET  & 0.72          & 0.71          & 0.77          & 0.66          & 0.72          & 0.66          \\  & MUSER    & 0.75          & 0.73          & 0.78          & 0.70          & 0.72          & 0.68          \\  & ReRead     & 0.71          & 0.71          & 0.75          & 0.68          & 0.71          & 0.69          \\ \hline
\multirow{5}{*}{\begin{tabular}[c]{@{}l@{}}LLM-based\\ methods\end{tabular}}                        & Vicuna-7B                                      & 0.52          & 0.53          & 0.52          & 0.51          & 0.52          & 0.51          \\     & WEBGLM                                         & 0.60          & 0.61          & 0.63          & 0.61          & 0.66          & 0.62          \\ & ProgramFC                                      & 0.73          & 0.72          & 0.74          & 0.63          & 0.62          & 0.64          \\ & STEEL                                          & 0.78          & 0.74          & 0.78          & 0.72          & 0.74          & 0.72          \\ \hline
\multicolumn{2}{l}{\textit{FakeCheckRAG}}                                                                                     & \textbf{0.85} & \textbf{0.83} & 0.86          & \textbf{0.83} & 0.84          & \textbf{0.83} \\ \hline

\multirow{3}{*}{Ablation Study}                                                                   & \textit{FakeCheckRAG}$_{3.5}$ & 0.82          & 0.71          & \textbf{0.99} & 0.74          & \textbf{0.98} & 0.59          \\ 
   & \textit{FakeCheck-NoRet}                                   & 0.57          & 0.55          & 0.56          & 0.55          & 0.56          & 0.57              \\  
 & \textit{FakeCheck-NoReS}                                 &  0.80             &    0.79           &     0.82          &   0.79            &0.80               &   0.78            \\ \hline
\end{tabular}
\end{table}

\begin{table}[htpb]
\centering
\footnotesize
\caption{
Comparison of our model's performance against baselines on LIAR dataset.}\label{tab:results2}
\begin{tabular}{llllllll}
\hline
\multicolumn{2}{l}{\multirow{2}{*}{Method}}                                                                                                      & \multicolumn{3}{c}{Real} & \multicolumn{3}{c}{Fake} \\ \cline{3-8} 
\multicolumn{2}{l}{}                                                                                                                             & F1     & P      & R      & F1     & P      & R      \\ \hline
\multirow{7}{*}{\begin{tabular}[c]{@{}l@{}}Classical\\ evidence\\ based\\ methods\end{tabular}} & DeClarE                                        & 0.53   & 0.55   & 0.54   & 0.61   & 0.58   & 0.59   \\  & HAN                                            & 0.56   & 0.54   & 0.53   & 0.60   & 0.61   & 0.61   \\      & EHIAN                                          & 0.55   & 0.54   & 0.54   & 0.63   & 0.60   & 0.61   \\  & MAC                                            & 0.56   & 0.55   & 0.56   & 0.62   & 0.62   & 0.62   \\  & GET                                            & 0.57   & 0.56   & 0.57   & 0.64   & 0.65   & 0.63   \\    & MUSER                                          & 0.64   & 0.64   & 0.65   & 0.64   & 0.65   & 0.63   \\ & ReRead                                         & 0.58   & 0.58   & 0.59   & 0.63   & 0.62   & 0.62   \\ \hline
\multirow{5}{*}{\begin{tabular}[c]{@{}l@{}}LLM-based\\ methods\end{tabular}}                  & Vicuna-7B                                      & 0.52   & 0.54   & 0.55   & 0.51   & 0.53   & 0.52   \\    & WEBGLM                                         & 0.55   & 0.56   & 0.57   & 0.62   & 0.60   & 0.61   \\  & ProgramFC                                      & 0.63   & 0.60   & 0.63   & 0.62   & 0.61   & 0.62   \\    & STEEL & 0.68   & 0.68   & 0.69   & 0.74   & 0.72   & 0.75   \\ \hline
                                        \multicolumn{2}{l}{\textit{FakeCheckRAG}}               &  \textbf{0.87}   & \textbf{0.83}   & 0.87   & \textbf{0.83}   & \textbf{0.84}   &\textbf{0.87}    \\ \hline
    \multirow{3}{*}{Ablation Study}                 
    & \textit{FakeCheckRAG}$_{3.5}$ & 0.80   & 0.68   & \textbf{0.89}   & 0.69   & 0.74   & 0.67                \\ 
   & \textit{FakeCheck-NoRet}                                   & 0.55   & 0.57   & 0.56   & 0.55   & 0.56   & 0.56   \\  
 & \textit{FakeCheck-NoReS}                                 &  0.78             &   0.74            &  0.82             & 0.71              &            0.72   &      0.69         \\ \hline
\end{tabular}
\end{table}
\subsubsection{Ablation Study}\label{sec:ablation}
To systematically evaluate the role and effectiveness of each component within the system, we performed an ablation study. Our complete system, \textit{FakeCheckRAG}, integrates RAG with GPT-4 and includes both a retrieval module and a re-search mechanism designed for iterative evidence collection. To isolate the impact of these components, we examined three ablated variants: \textit{FakeCheckRAG}$_{3.5}$, which replaces GPT-4 with GPT-3.5-turbo; \textit{FakeCheck-NoRet}, which disables the retrieval module; and \textit{FakeCheck-NoReS}, which omits the re-search process responsible for iterative querying. Finally, we assess the impact of iterative re-searching. Across both datasets, we observe consistent trends highlighting the importance of system components:

\textbf{Effect of Backbone Language Model (GPT-4 vs. GPT-3.5):} To isolate the effect of the underlying language model, we compare two variants of our system: \textit{FakeCheckRAG}, which utilizes GPT-4, and \textit{FakeCheckRAG}$_{3.5}$, which uses GPT-3.5-turbo. Both models share the same retrieval-augmented architecture and re-search mechanism, differing only in the language model component. On both PolitiFact and LIAR, \textit{FakeCheckRAG} consistently outperforms its GPT-3.5 counterpart. For instance, on PolitiFact, \textit{FakeCheckRAG} achieves an F1 of 0.85 and a precision of 0.83, compared to 0.82 and 0.71 for \textit{FakeCheckRAG}$_{3.5}$ for real news. A similar pattern is observed on LIAR, where the F1 improves from 0.80 to 0.87 and the precision from 0.68 to 0.83 on real claims. These results highlight the critical role of model capabilities in retrieval-augmented fact-checking systems.

A detailed analysis of the results from \textit{FakeCheckRAG}$_{3.5}$ reveals a critical insight: the GPT-3.5-turbo backbone induces a strong prediction skew that undermines reliability. On the PolitiFact dataset, \textit{FakeCheckRAG}$_{3.5}$ achieves a near-perfect Recall of 0.99 for real claims and a near-perfect Precision of 0.98 for fake claims. However, this reflects a severe imbalance. The model tends to classify claims as “real” (high Recall for real) and is very selective about “fake” (high Precision for fake). This strategy severely harms the complementary metrics: Precision for real is low (0.71), meaning it often labels fake claims as real, and Recall for fake is very low (0.59), meaning it fails to detect a large portion of fake claims. A similar, though less pronounced, pattern is observed on the LIAR dataset.
In contrast, \textit{FakeCheckRAG} system with GPT-4 demonstrates a more balanced reasoning capability. This leads to the superior overall F1 scores observed in our main results.

\textbf{Effect of Removing Retrieval:} The \textit{FakeCheck-NoRet} variant yields the lowest performance across all metrics on both datasets, demonstrating the critical importance of the retrieval component. Its removal leads to substantial degradation, confirming that access to external evidence is essential for accurate claim verification. The retrieval module plays a pivotal role by providing relevant evidence aligned with the input claim, enabling \textit{FakeCheckRAG} to make more informed and accurate predictions.

\textbf{Effect of Removing Re-Search:} The \textit{FakeCheck-NoReS} variant shows moderate performance degradation, lower than the full system but markedly better than \textit{FakeCheck-NoRet}. This suggests that the re-search mechanism contributes to improved evidence quality by enabling iterative retrieval, though its impact is secondary to the initial retrieval step. The performance gap highlights that a single-pass search often fails to provide sufficiently comprehensive evidence, underscoring the value of the re-search module in enhancing the system’s reasoning and prediction accuracy.

\begin{figure}[]
\centering
\includegraphics[width=0.6\textwidth]{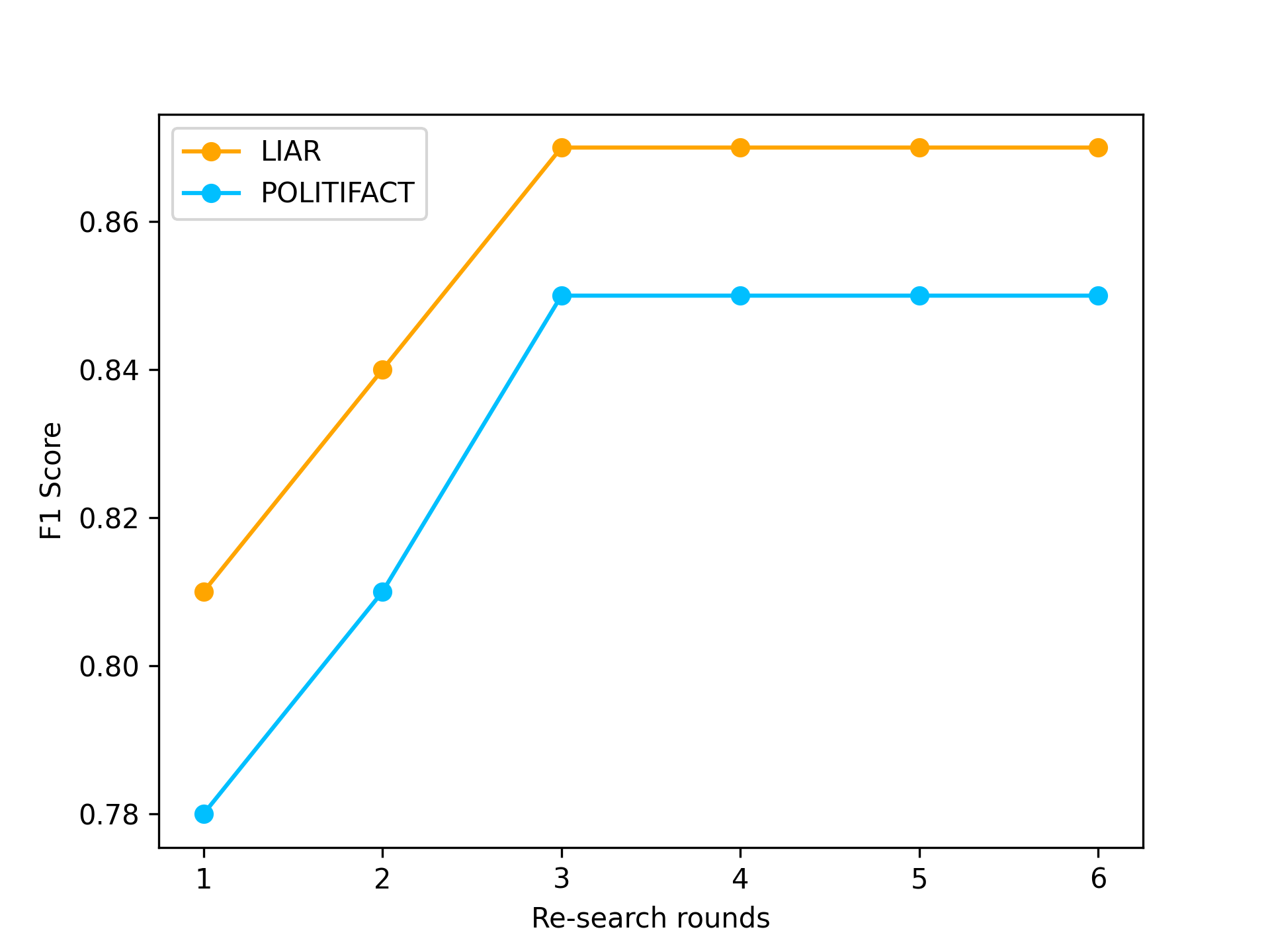}
\caption{F1 score across re-search rounds.}
\label{fig:researchRounds}
\end{figure}

\textbf{Effect of Re-Search Rounds:} To assess the impact of iterative re-searching on claim verification, we evaluate model performance across varying numbers of re-search rounds (from 1 to 6), as shown in Figure~\ref{fig:researchRounds}. 

On the PolitiFact dataset, the F1 score increases from approximately 0.78 with a single re-search round to around 0.85 by the third round, after which it stabilizes. A similar pattern is observed on the LIAR dataset, where F1 improves from about 0.81 to 0.87 by the third round, with minimal gains beyond that point. Notably, the optimal number of re-search rounds—three—remains consistent across both datasets. These findings suggest that iterative re-search contributes meaningfully to improved evidence acquisition and veracity detection. However, the marginal benefit diminishes after three rounds, indicating a performance plateau. Based on this observation, we adopt three re-search rounds as the default setting in our system.

These results indicate that incorporating multiple re-search rounds can enhance claim verification performance, especially in the earlier stages, though returns diminish beyond three rounds.

\textbf{Latency Analysis:}
To assess the practical deployability of \textit{FakeCheckRAG} as a browser extension, we measured end-to-end latency across various claims. Processing time varies significantly based on the claim, ranging from 8 seconds to 17 seconds. Latency is primarily influenced by three key factors: (1) claim complexity, where simple factual claims verify faster than complex claims requiring nuanced interpretation; (2) evidence volume, as claims with abundant search results require more processing time for evidence analysis; and (3) re-search necessity, since controversial claims often trigger multiple re-search rounds to gather contrasting perspectives. While 17 seconds is not instantaneous, we argue it represents a reasonable trade-off for fact-checking where accuracy is paramount. The iterative re-search mechanism, which contributes to this latency, is precisely what enables our significant accuracy improvements (15-20\% F1 gain over baselines). We also note that the Chrome extension can operate asynchronously - users can continue browsing while verification occurs in the background.

\section{User Study Evaluation}\label{sec:userstudy}
While achieving high classification accuracy is essential, the practical utility of \textit{Aletheia} also depends on how users perceive and interact with it. To explore this user-centered dimension, we conducted a user study in which participants engaged with our browser extension and completed a structured evaluation. Human evaluations were conducted over a two-month period, involving both in-lab participants from our research group and external participants recruited through SurveySwap\footnote{https://surveyswap.io/surveys?showModal=true?} and PollPool\footnote{https://www.poll-pool.com/app/dashboard} platforms. As both the survey and the \textit{Aletheia} interface were primarily in English, we selected participants with English proficiency from diverse global regions. A total of 250 individuals participated in the study and evaluated \textit{Aletheia}.

The evaluation followed a within-subjects design in which all participants completed each of the three rounds sequentially. In the first round, we collected demographic information as well as data on participants’ news consumption habits. In the second round, participants were introduced to \textit{Aletheia} through an explanatory video demonstrating its key features and how they function. To evaluate the usability and perceived effectiveness of \textit{Aletheia}’s three integrated components, participants were asked to respond to a series of Likert-scale questions on a 5-point scale ranging from 1 (Strongly Disagree) to 5 (Strongly Agree). The third round focused on system usability, using standardized methods to evaluate overall user experience.
\subsection{Participants characteristics}
The survey drew from a diverse group. In terms of age, the distribution was: 48\% were aged 25–34, followed by 40.4\% aged 18–24, 10.4\% aged 35–44, and 1.2\% aged 45–54. Regarding gender identity, 32.6\% of respondents identified as male, 35.6\% as female, 29.8\% preferred not to disclose, and 2\% identified as non-binary. Regarding profession, participants came from a wide range of sectors. The most common fields were Science/Research  (22.8\%), Technology (20.4\%), and Arts/Media (18.2\%). In terms of fake news exposure, 27.6\% reported rarely encountering it, while 38.8\% said they encountered it often. When asked about news sources, the most frequently cited were social media  (25.2\%) and online news websites  (24.0\%), followed by podcasts (23.6\%) and television (20.2\%). Participants' familiarity with fact-checking tools also varied: 26.2\% reported no familiarity, while 10.2\% were extremely familiar. Similarly, confidence in detecting fake news ranged widely, with 25.2\% not confident and 19.8\% extremely confident. Finally, news-sharing behavior on social media was: 30.4\% shared rarely, and 21.2\% frequently. Table~\ref{tab:userCharacteristics} provides a detailed breakdown of participant demographics.

\subsection{Component-wise Evaluation}
On average, participants rated all three components favorably: 
VerifyIt (mean $[M] = 4.00$, standard deviation $[SD] = 0.42$), Discussion Hub ($M = 3.98$, $SD = 0.38$) and Stay Informed ($M = 3.99$, $SD = 0.46$).  The small standard deviations indicate strong consensus and consistently positive evaluations across the participant pool. 

\begin{figure}[]
\centering
\includegraphics[width=\textwidth]{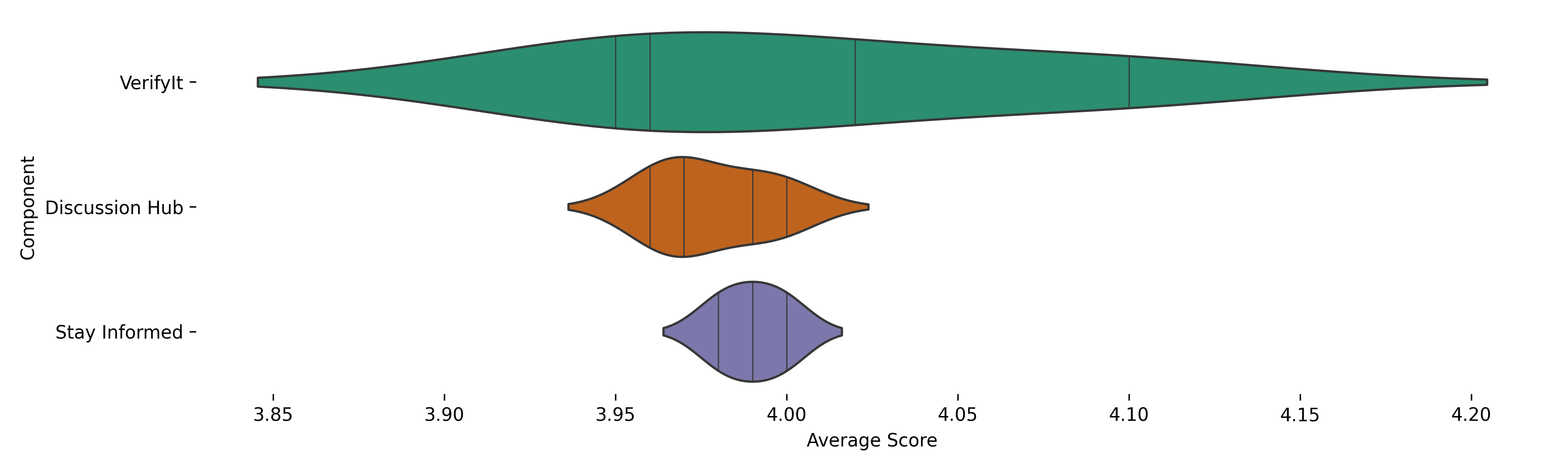}
\caption{Distribution of user ratings for three core components of \textit{Aletheia}.}
\label{fig:componenetWiseEvaluation}
\end{figure}

To further illustrate the distribution of user responses across components, Figure~\ref{fig:componenetWiseEvaluation} presents a violin plot of average ratings. \textit{VerifyIt} exhibited the highest median rating and the widest distribution, indicating strong user approval alongside diverse experiences. In contrast, \textit{Discussion Hub} and \textit{Stay Informed} showed narrower distributions centered around slightly lower medians, suggesting more consistent but comparatively moderate perceptions of usefulness and engagement.

A breakdown of individual item scores for each component is presented in Table~\ref{tab:component_feedback}, offering a more granular view of user responses across specific functionality areas.

\begin{table}[]
\caption{Mean user ratings (1–5) for individual Likert-scale items assessing the core components of \textit{Aletheia}: VerifyIt, Discussion Hub, and Stay Informed.}
\label{tab:component_feedback}
\begin{tabular}{cp{9cm}c}
\hline
\textbf{Component}              & \textbf{Question}                                                                                  & \textbf{Mean} \\ \hline
\multirow{4}{*}{VerifyIt}       & I found Verify It's fact-checking process transparent and understandable.                          & 3.95                   \\ \cline{2-3} 
                                & I would trust Verify It to help me make informed decisions about news content.                     & 3.96                   \\ \cline{2-3} 
                                & Verify It's explanations help me understand why a news piece was flagged as false or misleading.   & 4.10                   \\ \cline{2-3} 
                                & I would recommend Verify It to others who are concerned about fake news.                           & 4.02                   \\ \hline
\multirow{5}{*}{\begin{tabular}[c]{@{}c@{}}Discussion\\ Hub\end{tabular}} & The Discussion Hub allow me to engage in meaningful conversations about news content.              & 3.97                   \\ \cline{2-3} 
                                & I found the user discussions helpful in assessing the credibility of news articles.                & 3.96                   \\ \cline{2-3} 
                                & The ability to see different perspectives in the Discussion Hub improves my understanding of news. & 4.00                   \\ \cline{2-3} 
                                & I feel comfortable participating in discussions in the Discussion Hub.                             & 3.99                   \\ \cline{2-3} 
                                & The Discussion Hub is a valuable addition to the fake news detection process.                      & 3.97                   \\ \hline
\multirow{3}{*}{\begin{tabular}[c]{@{}c@{}}Stay \\ Informed\end{tabular}}  & The Stay Informed section helps me stay updated on the latest fact-checks.                         & 3.98                   \\ \cline{2-3} 
                                & I found the Stay Informed section useful for identifying trending fake news.                       & 4.00                   \\ \cline{2-3} 
                                & The Stay Informed feature encourages me to fact-check news more often.                             & 3.99                   \\ \hline
\end{tabular}
\end{table}

\subsubsection{VerifyIt}
The VerifyIt component received consistently favorable evaluations, reflecting strong user trust in its explanatory capabilities. The highest-rated item, \textit{``VerifyIt’s explanations help me understand why a news piece was flagged as false or misleading''} ($M = 4.10$), underscores users’ appreciation for clarity and reasoning rather than opaque verdicts. This suggests that transparency in automated fact-checking—particularly the ``why'' behind a label—is key to fostering comprehension and trust.

This preference for interpretability was echoed in open-ended feedback. One participant remarked, \textit{``I like the way it would allow me to detect fake news and understand why they are fake,''} while another praised the integration of AI and evidence: \textit{``I liked how Aletheia combines AI detection with explanations and evidence from credible sources.''} These comments emphasize that users do not see VerifyIt merely as a filter but as a learning tool—one that helps build their capacity for independent judgment.

Other positively rated items, such as trust in VerifyIt’s guidance ($M = 3.96$) and willingness to recommend it to others ($M = 4.02$), further suggest that the component successfully bridges usability and credibility. The fact-checking process was also seen as transparent and understandable ($M = 3.95$), reinforcing the notion that the system’s design aligns with user expectations for openness and reliability.

\subsubsection{Discussion Hub}
The Discussion Hub revealed a more nuanced picture. While participants acknowledged the benefits of engaging with diverse perspectives—reflected in high ratings such as \textit{``The ability to see different perspectives improves my understanding of news''} ($M = 4.00$)—there was also evidence of cautious optimism. Scores indicating user comfort with participation ($M = 3.99$) and the perceived value of the Hub in the fact-checking process ($M = 3.97$) show that users generally accepted the idea of social deliberation, though not uncritically.

Open-ended feedback surfaced key concerns that mirror broader anxieties about online discourse. One participant noted, \textit{``The discussion hub could be vulnerable to adversarial users,''} while another warned, \textit{``I can imagine people deploying bots to spread fake news.''} These comments point to a tension between the ideal of collaborative sensemaking and the reality of adversarial digital environments.

Nonetheless, the fact that many users still found the space helpful in assessing article credibility ($M = 3.96$) and engaging in meaningful conversations ($M = 3.97$) highlights the feature’s potential. Designing mechanisms that support respectful dialogue and mitigate manipulation will be essential to realizing this potential more fully.

\subsubsection{Stay Informed}
The Stay Informed component was well-received for its role in promoting ongoing awareness. Users rated its usefulness in identifying trending fake news highly ($M = 4.00$), with similar scores for its ability to keep them updated ($M = 3.98$) and encourage fact-checking behavior ($M = 3.99$). These ratings indicate that users did not just find the feature informative—they found it motivating.

Participants described the component as reinforcing habitual engagement. One stated, \textit{``The Stay Informed feature is very good and as you said, it encourages me to fact-check news more often,''} highlighting the subtle nudging effect that this kind of information feed can provide. Rather than overwhelming users with alerts, Stay Informed seems to function as a quiet but persistent reminder to remain vigilant. In this sense, Stay Informed acts less like a reactive alert system and more like a behavioral scaffold—helping users integrate fact-checking into their broader media consumption routines. Its value lies not only in the information it provides but in its capacity to promote sustained attention to misinformation trends over time.

\subsubsection{Perceived Balance and Cross-User Consistency}
To assess whether users perceived any one component as significantly more useful than the others, we conducted a repeated measures ANOVA. The analysis revealed no significant differences in ratings across the three components, $F$(2, 498) = 0.29, $p$ = 0.75, $\eta^2_g$ = 0.00076. Post-hoc pairwise comparisons using Bonferroni correction confirmed that all differences were negligible (all $p$ = 1.00, Cohen’s $d$ < 0.08). These findings suggest that users regarded the three components as comparably useful, supporting the interpretation that \textit{Aletheia}’s design is functionally well-balanced—no single feature was seen as disproportionately strong or weak.

To explore whether evaluations varied across different user groups, we conducted a series of moderation analyses incorporating demographic and behavioral factors (e.g., age, gender, news consumption habits, and prior experience with fact-checking tools). No significant interaction effects emerged, indicating that perceptions of each component’s usefulness were consistent across participant subgroups. This suggests that the system’s appeal and usability were broadly inclusive, resonating similarly across diverse user backgrounds. For full statistical details, see Appendix~\ref{tab:interaction-effects}.

\subsection{System Usability Evaluation}
In this study, usability is defined as the overall quality of the user experience (UX) when interacting with \textit{Aletheia}. This evaluation focuses on users’ subjective satisfaction with the system’s interface, ease of use, and interaction design. It does not assess the system’s effectiveness in identifying fake news.

To measure usability, we employed the System Usability Scale (SUS), a widely used instrument in human-computer interaction research~\citep{putra2024evaluation}. The SUS consists of 10 standardized items rated on a 5-point Likert scale ranging from  ``strongly disagree'' to ``strongly agree,'' providing a concise and reliable assessment of perceived usability.

For this study, the original SUS items were slightly adapted to reflect the context of interacting with a fake news detection system. These adaptations preserved the semantic integrity of each item to maintain the scale’s psychometric validity. The adapted items are listed below:
\begin{enumerate}
\item I would use \textit{Aletheia} frequently if integrated into my browser.
\item I found \textit{Aletheia} unnecessarily complex.
\item \textit{Aletheia} was easy to use, and I quickly learned how to navigate its features.
\item I would need technical support to use \textit{Aletheia} effectively.
\item The components (VerifyIt, Discussion Hub, Stay Informed) are well-integrated.
\item I noticed inconsistencies in \textit{Aletheia}’s interface or functionality.
\item Most people would learn to use \textit{Aletheia} quickly.
\item \textit{Aletheia} felt cumbersome to interact with.
\item I felt confident using \textit{Aletheia} to evaluate news.
\item I needed to learn many steps before using \textit{Aletheia} effectively.
\end{enumerate}
The overall SUS score was calculated using the standard scoring method:
\begin{equation}
SUS = \frac{1}{n} \sum_{i=1}^{n} \left[ \left( \sum_{j=1,3,5,7,9} (x_{ij} - 1) + \sum_{j=2,4,6,8,10} (5 - x_{ij}) \right) \times 2.5 \right]
\label{eq:combined_sus}
\end{equation}
where \( x_{ij} \) denotes the score given by respondent \( i \) to question \( j \).

We interpreted the final SUS scores using a standard grading scale~\citep{putra2024evaluation}. Scores above 85 are classified as \textit{Excellent}, 70–84 as \textit{Good}, 51–69 as \textit{OK}, and 0–50 as \textit{Poor}, reflecting increasing levels of usability concern.

Across the sample of 250 participants, \textit{Aletheia} achieved a mean SUS score of 83.4, placing it in the \textit{Good} category. This indicates that users found the extension easy to use, with features that were well-integrated and learnable. The standard deviation ($SD = 4.2$) was relatively low, indicating strong agreement among participants and consistent positive experiences.

This finding is particularly notable in a domain where few misinformation tools have undergone formal usability evaluation. Furthermore, the high SUS score aligns with component-level results, where each major feature received average ratings near 4.0 on a 5-point scale. This convergence confirms the platform’s overall design strength and user-centered coherence.

To provide a more detailed view of usability, we analyzed individual SUS item responses, examined the distribution of overall scores, and segmented results by user subgroups. Item-level analysis showed strong agreement with positively worded items—especially those related to ease of use, confidence, and learnability—and low agreement with negatively worded items assessing inconsistency, complexity, or reliance on support. These results suggest that participants found \textit{Aletheia} to be intuitive, coherent, and easy to navigate.

The distribution of SUS scores was positively skewed, with most ratings clustering between 75 and 90, indicating widespread satisfaction. Additionally, subgroup analyses showed consistent SUS ratings across key demographic and behavioral categories (e.g., gender, age, fake news exposure, and news-sharing frequency). Only minor variations were observed, suggesting that perceptions of usability were broadly robust across diverse user profiles. For full item-level scores, score distributions, and subgroup comparisons, see~\ref{appendix:sus}.
\section{Discussion, Limitations, and Future Work}\label{secDiscuss}
The dual-layer evaluation of \textit{Aletheia}, spanning both model performance and user-centered feedback, provides a comprehensive assessment of its effectiveness as a fake news detection tool.

The VerifyIt component, powered by the FakeCheckRAG model, consistently outperformed both classical and LLM-based baselines across two benchmark datasets. Importantly, this technical superiority was mirrored by users’ perceptions. VerifyIt received the highest ratings among all components, with participants commending its explanation quality, transparency, and decision support. The ability to understand \textit{why} a claim was flagged as false was repeatedly highlighted as a critical feature, reinforcing the value of interpretable AI in fact-checking applications.

Nonetheless, the current system employs a static blacklist during evidence preprocessing to filter out known disinformation sources. While effective to a degree, this static approach may be inadequate given the scale, diversity, and rapid evolution of digital content. Future iterations should explore dynamic or model-driven mechanisms for detecting and excluding untrustworthy sources, potentially integrating source credibility assessments directly into the retrieval process. As a future direction, we will treat the blacklist as an \textit{initial seed} that can be automatically augmented over time (e.g., via user feedback and cross-source consistency signals), while the present results use the static setting for strict comparability.
 Moreover, given that fake news can be mass-produced and distributed across multiple articles, the ability to verify clusters of related claims—rather than isolated statements—may significantly enhance detection capabilities.

In addition, the system can process claims and evidence across languages without architectural changes, leveraging GPT-4's training on diverse multilingual corpora~\citep{yoo2025multilingual}. While our evaluation focused on English, future work will quantitatively benchmark performance across languages and integrate region-specific credibility datasets. Regionally, we surface locally relevant sources by issuing locale-aware queries via Google Custom Search Engine, apply region-specific credibility lists in addition to the global blacklist, and harmonize regional conventions (dates, numbers, units); future work will systematically evaluate performance across regions and resolve cross-region source conflicts.

Effectiveness also hinges on how users interact with the system’s broader ecosystem. The Discussion Hub was positively received for fostering diverse perspectives and deliberative engagement. However, participants also raised concerns about the potential for adversarial behavior, misinformation propagation, and unmoderated discourse. These insights underscore a key limitation: while engagement features enrich the fact-checking experience, they must also be designed to withstand manipulation. Addressing this tension will require the implementation of trust scoring, moderation tools, or community flagging mechanisms to safeguard the integrity of user interactions.

The Stay Informed component was similarly well-regarded. Participants noted that it served as a subtle but persistent prompt to remain attentive to emerging misinformation trends. Rather than functioning as a reactive alert system, this feature encouraged proactive, habitual engagement with fact-checking resources. Such behavioral nudging may play an important role in fostering long-term media literacy and critical consumption practices.

Despite these promising findings, several limitations must be acknowledged. First, the participant pool was self-selected and recruited primarily through online platforms, which may limit generalizability to broader populations. The study was also conducted exclusively in English, restricting applicability across linguistic and regional contexts. Second, the evaluation captured short-term impressions based on structured interactions with the system. As such, it does not reflect long-term usage patterns, shifts in trust, or sustained behavioral change.

Future research should include longitudinal studies to assess the enduring impact of \textit{Aletheia} on user habits, comprehension, and decision-making. Expanding language coverage and introducing safeguards in user-generated discussion spaces are also critical priorities. Despite these limitations, the consistently high ratings across components and the strong SUS score indicate that \textit{Aletheia} is a promising and user-friendly tool. Moving forward, the focus will be on scaling its capabilities, enhancing resilience against misuse, and deepening its role in supporting informed news consumption in diverse real-world contexts.
\section{Conclusion}\label{secConclusion}
This paper presented \textit{Aletheia}, a fake news detection browser extension. Our evaluation showed that the system achieves strong detection performance while also being perceived as transparent, usable, and trustworthy by users. The VerifyIt component outperformed competitive baselines while providing transparent, evidence-backed explanations. Complementary features, Discussion Hub and Stay Informed, were positively received, highlighting the value of social deliberation and behavioral reinforcement in combating fake news. While challenges remain, particularly regarding adversarial risks, multilingual support, and long-term user engagement, our findings suggest that integrating AI with thoughtful interaction design holds substantial promise. Future work will focus on enhancing robustness, expanding inclusivity, and evaluating the longitudinal impact of such systems in real-world settings.






\bibliographystyle{cas-model2-names}

\bibliography{cas-refs}
\newpage
\appendix               
\section{Prompts Used}
\begin{longtable}{lp{13cm}}
\caption{Prompts used.} \label{tab:agent-prompts} \\
\toprule
\textbf{Step} & \textbf{Prompt} \\
\midrule
\endfirsthead
\toprule
\textbf{Step} & \textbf{Prompt} \\
\midrule
\endhead
\textbf{\begin{tabular}[c]{@{}l@{}}Claim\\ Extraction\end{tabular}}  &
\begin{minipage}[t]{\linewidth}
\texttt{Given the input content below, please summarize the single key claim.} \\
\texttt{Input content: \{content\}} \\
\texttt{Please output with the following JSON format: \{"key\_claim": "XXX"\}}
\end{minipage} \\
\hline
\textbf{\begin{tabular}[c]{@{}l@{}}Query \\ Generation\end{tabular}}  &
\begin{minipage}[t]{\linewidth}
\texttt{Given the claim below, please generate a Google query which} \\
\texttt{can be used to search content to verify this claim.} \\
\texttt{Claim: \{claim\}} \\
\texttt{Please output with the following JSON format: \{"query": "XXX"\}}
\end{minipage}  \\
\hline
\textbf{\begin{tabular}[c]{@{}l@{}}Web \\ Search\end{tabular}}  & \begin{minipage}[t]{\linewidth}No prompt, this component performs external retrieval via the Google  Search API based on the generated query.\end{minipage} \\
\hline
\textbf{\begin{tabular}[c]{@{}l@{}}Evidence \\ Evaluation\end{tabular}}  &
\begin{minipage}[t]{\linewidth}
\texttt{Below is one web search result:} \\
\texttt{    Search Result:  \{search\_result\}} \\
\texttt{    Below is a claim to be verified:  Claim: \{claim\}} \\
\texttt{    Please perform the following rules to generate an output with this JSON format:} \\
\texttt{    \{"support\_or\_contradict\_or\_unrelated": "support" or "contradict" or "unrelated", "confidence": XX (0-100), "rationale": "XXX"\}} \\
\texttt{ Rule 1: if the search result content supports the claim,} \\
\texttt{set the field as "support", and offer a confidence score and rationale.} \\
\texttt{Rule 2: if the content negates the claim, set the field as "negate".} \\
\texttt{Rule 3: if it cannot support or negate, use "baseless".} \\
\texttt{To clarify: if the result does not contradict the claim,} \\
\texttt{but lacks supporting info, use "baseless" rather than "negate".}
\end{minipage} \\ 
\hline
\textbf{Decision}  &
\begin{minipage}[t]{\linewidth}
\texttt{You are an assistant that determines the veracity of a claim based on multiple} \\ \texttt{pieces of evidence. Claim: \{claim\}} \\
\texttt{Evidence and Analysis: \{analyses\_text\}} \\
\texttt{Based on the provided web search results, analyze whether the information} \\
\texttt{has enough evidence to decide whether the statement is real or fake.} \\
\texttt{- If you conclude the statement is true, classify it as "real".} \\
\texttt{- If you conclude the statement is false, classify it as "fake".} \\
\texttt{- If the evidence is mixed or insufficient to make a determination,} \\
\texttt{  classify it as "NEI" (Not Enough Information).} \\
\texttt{Provide your answer in the following JSON format:} \\
\texttt{\{} 
\texttt{\ \ \ \ "decision": "real" or "fake" or "NEI",} \\
\texttt{\ \ \ \ "confidence": XX  \# Confidence score as a percentage between 0 and 100} 
\texttt{\}}
\end{minipage} \\
\hline
\textbf{Explanation}  &
\begin{minipage}[t]{\linewidth}
\texttt{You are an assistant that generates an explanation for a decision based solely} \\
\texttt{on the text of the claim and the classification.} \\
\texttt{Claim: \{claim\}} \\
\texttt{Decision: \{decision\}} \\
\texttt{Confidence:\{confidence\}\%} \\
\texttt{Based on the claim and the decision, provide a detailed explanation for the classification.} \\
\texttt{The explanation should include reasoning behind the decision,} \\
\texttt{including any relevant context that could support the decision.} \\
\texttt{If the decision is "real" or "fake", explain why.} \\
\texttt{Provide your answer in the following JSON format:} \\
\texttt{\{ "explanation": "<explanation text>" \}}
\end{minipage}  \\
\bottomrule
\end{longtable}

\section{User Study Sample Characteristics}
\begin{table}[H]
\centering
\caption{Detailed Breakdown of Survey Respondent Characteristics}\label{tab:userCharacteristics}
\footnotesize
\begin{tabular}{p{4cm} p{4.5cm} r r}
\toprule
\textbf{Category} & \textbf{Response} & \textbf{Count} & \textbf{Percentage} \\
\midrule
\multirow{4}{*}{Age}
  & 18–24 & 101 & 40.4\% \\
  & 25–34 & 120 & 48.0\% \\
  & 35–44 & 26 & 10.4\% \\
  & 45–54 & 3 & 1.2\% \\
\midrule
\multirow{4}{*}{Gender}
  & Male & 82 & 32.6\% \\
  & Female & 89 & 35.6\% \\
  & Non-binary & 5 & 2.0\% \\
  & Prefer not to say & 74 & 29.8\% \\
\midrule
\multirow{7}{*}{Professional Field}
  & Education & 47 & 18.8\% \\
  & Technology & 51 & 20.4\% \\
  & Healthcare & 22 & 8.6\% \\
  & Arts/Media & 46 & 18.2\% \\
  & Science/Research & 57 & 22.8\% \\
  & Business & 16 & 6.4\% \\
  & Other & 11 & 4.8\% \\
\midrule
\multirow{4}{*}{\begin{tabular}[c]{@{}l@{}}Frequency of \\ Fake News \\ Exposure\end{tabular}}
  & Never & 19 & 7.6\% \\
  & Rarely & 69 & 27.6\% \\
  & Sometimes & 65 & 26.0\% \\
  & Often & 97 & 38.8\% \\
\midrule
\multirow{5}{*}{Primary News Source}
  & Social media platforms  & 63 & 25.2\% \\
  &Online news websites  & 60 & 24.0\% \\
  & Podcasts & 59 & 23.6\% \\
  & Television & 51 & 20.2\% \\
  & Print newspapers  & 17 & 7.0\% \\
\midrule
\multirow{5}{*}{\begin{tabular}[c]{@{}l@{}}Familiarity with \\ Fact-Checking Tools\end{tabular}}
  & Not familiar & 66 & 26.2\% \\
  & Slightly familiar & 50 & 20.0\% \\
  & Moderately familiar & 61 & 24.4\% \\
  & Very familiar & 48 & 19.2\% \\
  & Extremely familiar & 25 & 10.2\% \\
\midrule
\multirow{5}{*}{\begin{tabular}[c]{@{}l@{}}Confidence in \\ Identifying Fake \\ News\end{tabular}}
  & Not confident & 63 & 25.2\% \\
  & Slightly confident & 52 & 20.6\% \\
  & Moderately confident & 42 & 16.8\% \\
  & Very confident & 44 & 17.6\% \\
  & Extremely confident & 49 & 19.8\% \\
\midrule
\multirow{4}{*}{\begin{tabular}[c]{@{}l@{}}News Sharing on \\ Social Media\end{tabular}}
  & Never & 48 & 19.0\% \\
  & Rarely & 75 & 30.4\% \\
  & Occasionally & 74 & 29.4\% \\
  & Frequently & 53 & 21.2\% \\
\bottomrule
\end{tabular}
\end{table}

\section{User Assessment of System Features}
\begin{table}[H]
\centering
\caption{
Mixed ANOVA interaction effects between component ratings and participant characteristics. 
None of the interactions were statistically significant ($p > 0.15$), and all partial eta-squared 
($\eta_p^2$) values indicate negligible practical effects. 
\textbf{F} denotes the F-statistic from the ANOVA test, 
\textbf{$p$-value} indicates the significance level, and 
\textbf{$\eta_p^2$} represents partial eta-squared, a measure of effect size.
}
\begin{tabular}{lccc}
\toprule
\textbf{Moderator} & \textbf{F} & \textbf{$p$-value} & \textbf{$\eta_p^2$} \\
\midrule
How often do you encounter fake news?         & 1.56 & 0.155 & 0.0187 \\
What are your primary news sources?           & 1.15 & 0.330 & 0.0184 \\
How familiar are you with fact-checking tools?& 0.83 & 0.572 & 0.0134 \\
How often do you share news on social media?  & 0.72 & 0.634 & 0.0087 \\
What is your age range?                       & 0.47 & 0.831 & 0.0057 \\
What is your gender?                          & 0.44 & 0.851 & 0.0054 \\
\bottomrule
\end{tabular}
\label{tab:interaction-effects}
\end{table}

\section{System Usability Scale — Detailed Evaluation Results}
\label{appendix:sus}

\begin{figure}[]
\centering
\includegraphics[width=\textwidth]{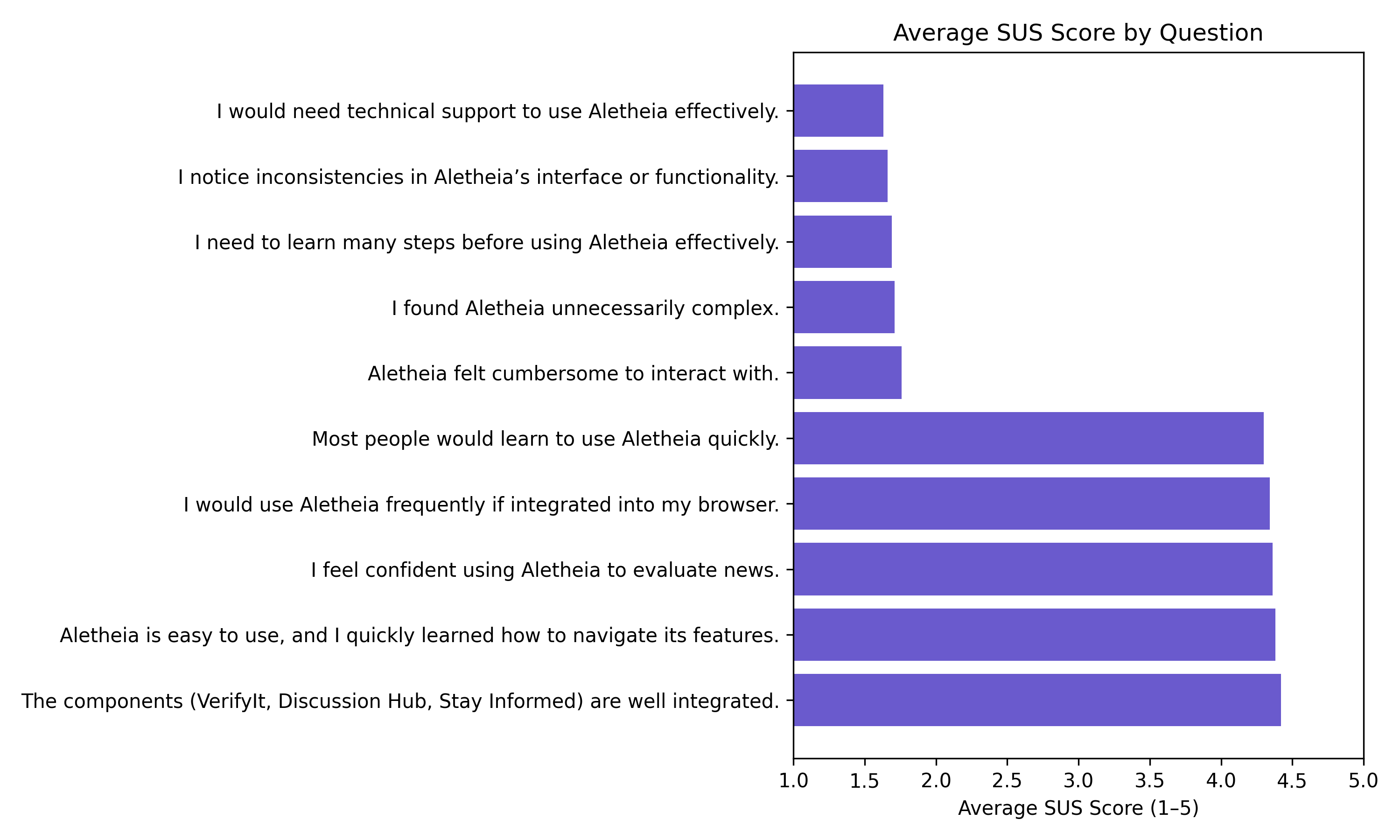}
\caption{Average scores for each of the ten individual items from the System Usability Scale, based on responses from all survey participants.}
\label{fig:susscoreperquestion}
\end{figure}

Each item was rated on a 5-point Likert scale (1 = Strongly Disagree, 5 = Strongly Agree), alternating between positively and negatively worded statements.
High scores on positively worded items (e.g., ease of use, confidence, quick learnability) and low scores on negatively worded items (e.g., perceived complexity, inconsistency, need for support) suggest that users found the system intuitive, consistent, and easy to learn (Figure~\ref{fig:susscoreperquestion}). This item-level pattern complements the overall SUS score and helps pinpoint specific usability strengths, supporting the conclusion that \textit{Aletheia} demonstrates a high level of perceived usability.

\begin{figure}[]
\centering
\includegraphics[width=\textwidth]{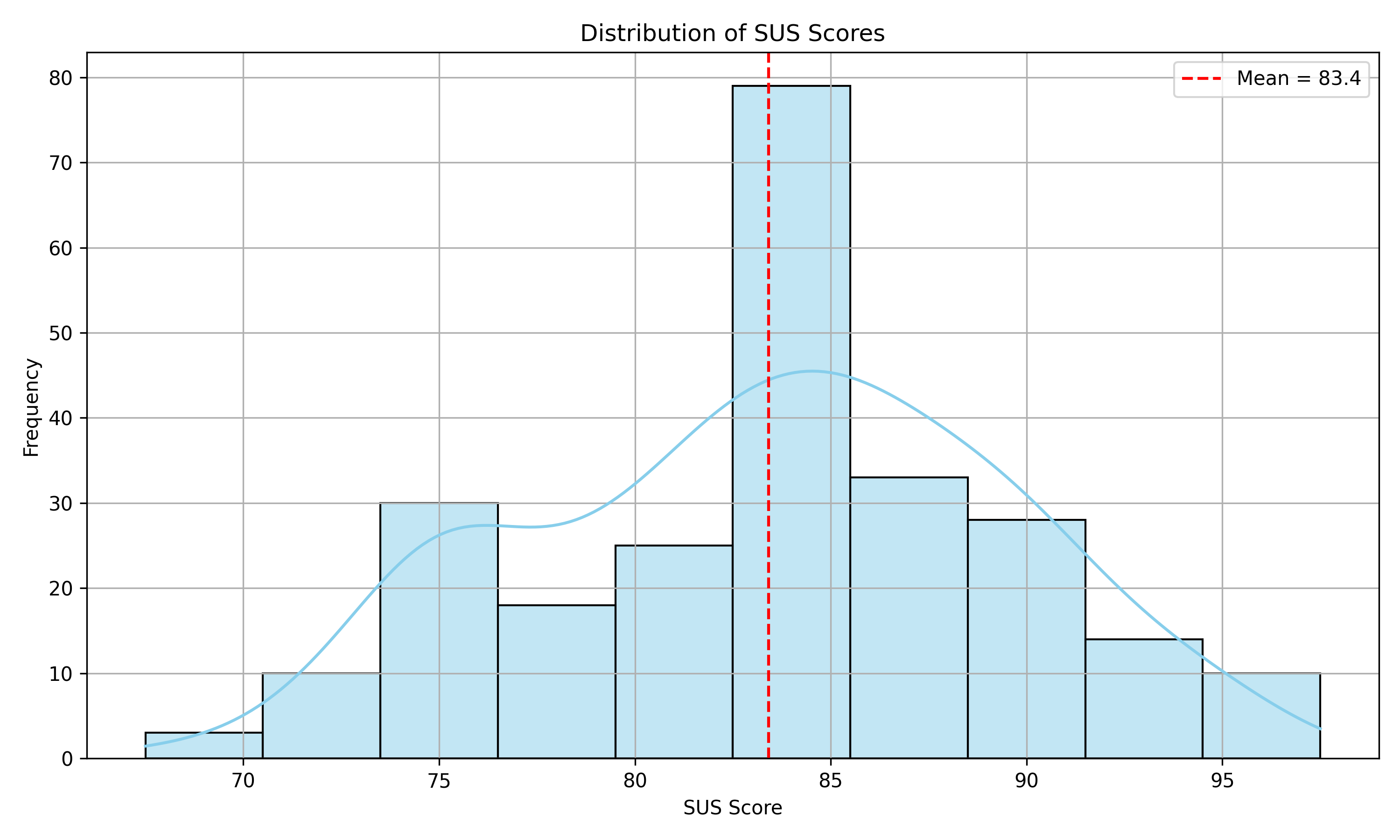}
\caption{Distribution of overall System Usability Scale scores reported by participants. Each SUS score represents a standardized usability rating derived from 10 Likert-scale items and scaled from 0 to 100. }
\label{fig:susdistribution}
\end{figure}

Figure~\ref{fig:susdistribution} illustrates the frequency of user scores across intervals, while the smooth blue kernel density estimation (KDE) curve estimates the underlying distribution. A red dashed vertical line indicates the mean SUS score of 83.4. The results show a concentration of scores between 75 and 90, with relatively few low values, indicating a positively skewed perception of usability. This suggests that \textit{Aletheia} was perceived not only as usable but also favorably consistent and intuitive across the participant sample.

\begin{figure}[]
\centering
\includegraphics[width=\textwidth]{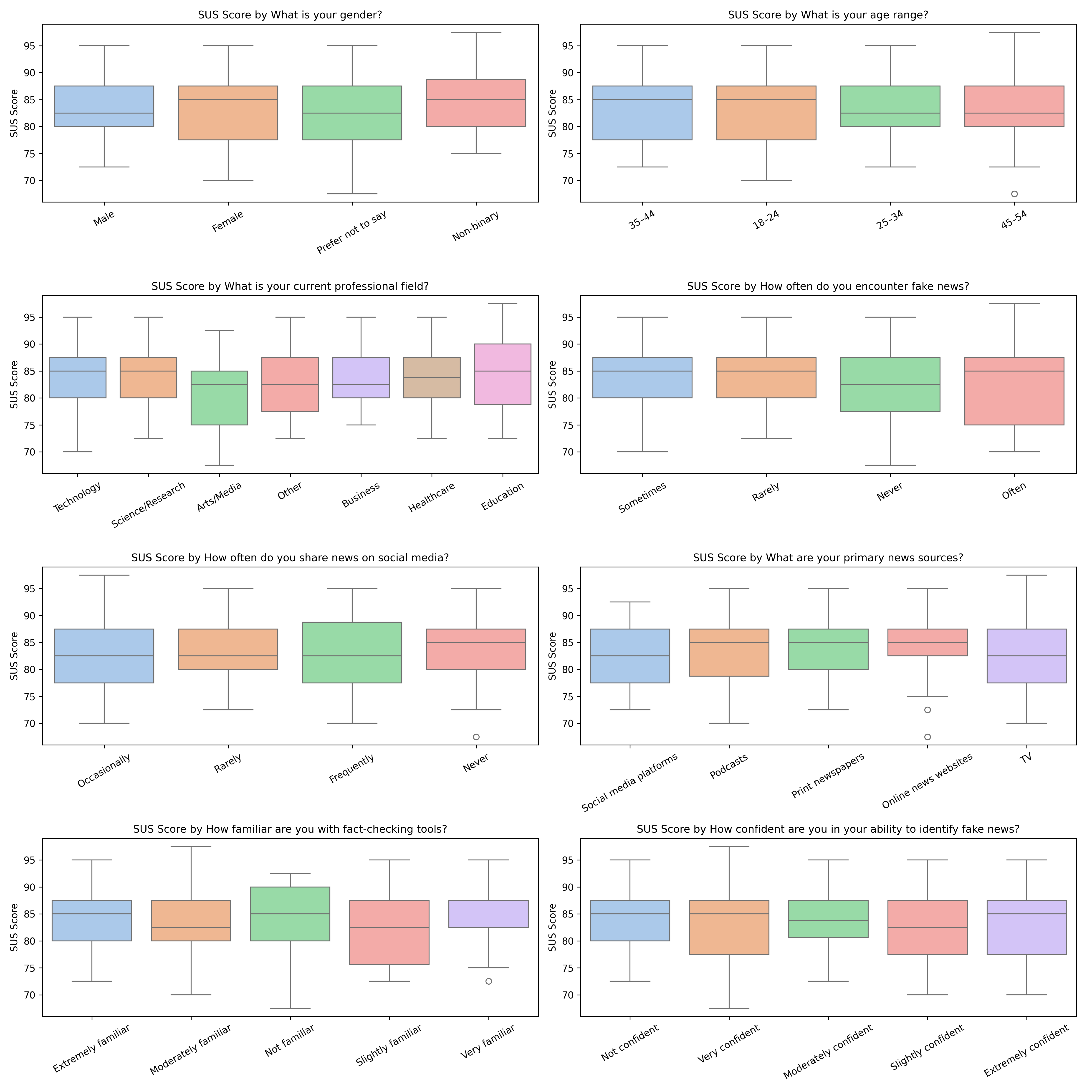}
\caption{Boxplots of System Usability Scale scores segmented by eight key respondent characteristics: gender, age range, professional field, frequency of fake news exposure, frequency of news sharing on social media, primary news source, familiarity with fact-checking tools, and confidence in identifying fake news.}
\label{fig:suspergroup}
\end{figure}

Each subplot, in Figure~\ref{fig:suspergroup}, illustrates the distribution of SUS scores for subgroups within each variable, with medians, interquartile ranges, and outliers displayed. Across all demographic and behavioral segments, SUS scores are relatively consistent, with medians generally clustered between 80 and 85. Notable observations include slightly higher median scores for non-binary respondents, educators, and those extremely familiar with fact-checking tools. Outliers (shown as dots or circles) appear sporadically, such as among respondents who report being “Very familiar” with fact-checking tools or “Never” sharing news, indicating isolated, unusually high or low usability perceptions. These results suggest that the usability of \textit{Aletheia} is broadly robust across diverse user groups, with no major usability concerns concentrated in any specific demographic or behavioral category.

\printcredits





\end{document}